\pdfoutput=1
\documentclass[11pt, letterpaper, onecolumn, copyright, numbering, logo]{paper_improved}

\usepackage[export]{adjustbox}
\usepackage{multirow}
\usepackage{float}
\usepackage{xcolor}
\usepackage{natbib}
\let\cite\citep

\usepackage[utf8]{inputenc} 
\usepackage[T1]{fontenc}    
\usepackage[most]{tcolorbox}

\usepackage{hyperref}       
\usepackage{url}            
\usepackage{booktabs}       
\usepackage{amsfonts}       
\usepackage{nicefrac}       
\usepackage{xcolor}         
\usepackage{graphicx}

\usepackage{natbib}
\usepackage{verbatim}
\usepackage{inconsolata}
\usepackage{tabularx}
\usepackage{float}
\usepackage{multirow}
\usepackage{listings}
\usepackage{xcolor}
\usepackage{svg}
\usepackage[T1]{fontenc}
\usepackage{wrapfig}

\usepackage{times}
\usepackage{url}
\usepackage{wrapfig,lipsum}
\usepackage{graphicx}
\usepackage{color}
\usepackage{colortbl}
\usepackage{booktabs} 
\usepackage{multirow}
\usepackage{array}
\usepackage[linesnumbered,ruled,vlined]{algorithm2e}
\usepackage{amsfonts}       
\usepackage{xcolor}         
\usepackage{float}
\usepackage{caption}
\usepackage[table]{xcolor}
\newcolumntype{C}[1]{>{\centering\arraybackslash}p{#1}}

\usepackage{float}
\usepackage{ragged2e}  
\usepackage{subcaption}
\usepackage{tabularx}
\usepackage[most]{tcolorbox}
\usepackage{listings}
\usepackage{tcolorbox}

\newtcolorbox[%
    auto counter,          
    number within=section, 
    list inside=promptlist 
]{promptbox}[1][]{
  colback=blue!5!white,
  colframe=blue!75!black,
  title=Decision Rules,
  fonttitle=\bfseries,
  boxrule=0.8mm,
  arc=2mm,
  #1 
}

\definecolor{lightgray}{rgb}{0.95, 0.95, 0.95}
\definecolor{darkgray}{rgb}{0.4, 0.4, 0.4}
\definecolor{backcolour}{rgb}{0.95,0.95,0.92}
\definecolor{myblue}{rgb}{0.2, 0.4, 0.8} 
\definecolor{mygreen}{rgb}{0.2, 0.6, 0.2} 

\usepackage{amsmath,amssymb,amsfonts}
\usepackage{soul}
\usepackage{url}
\usepackage[utf8]{inputenc}
\usepackage{caption}
\usepackage{graphicx}
\usepackage{xcolor}

\usepackage{amsthm}
\usepackage{booktabs}
\usepackage{latexsym}
\usepackage{graphicx} 
\usepackage{microtype}
\usepackage[switch]{lineno}
\usepackage{alltt}

\usepackage[table]{xcolor} 

\definecolor{mybgcolor}{HTML}{E6F2FF}

\definecolor{forestgreen}{rgb}{0.13, 0.55, 0.13}

\tcbset{
    aibox/.style={
        colback=white, 
        colframe=blue!75!black, 
        colbacktitle=blue!85!black, 
        coltitle=white, 
        fonttitle=\bfseries,
        enhanced,  
        drop shadow=black!50!white,  
        boxrule=1pt,  
        boxsep=10pt,  
        left=10pt,  
        right=10pt,  
        top=6pt,  
        bottom=6pt,  
        title code={\path[tcb fill frame] ([xshift=-10pt]frame.west) -- (frame.north west) -- (frame.north east) -- ([xshift=10pt]frame.east) -- cycle;},  
        attach boxed title to top left={xshift=10pt, yshift*=-\tcboxedtitleheight/2},
        boxed title style={boxrule=0pt, frame code={}}  
    }
}

\newtcolorbox{AIbox}[2][]{aibox, title=#2, #1}
\let\cite\citep
\title{Marco DeepResearch: Unlocking Efficient Deep Research Agents
via Verification-Centric Design}

\author[*,1]{Bin Zhu$^\dagger$, Qianghuai Jia$^\dagger$, Tian Lan$^\dagger$, Junyang Ren$^\dagger$, Feng Gu$^\dagger$, Feihu Jiang$^\dagger$, Longyue Wang{$^{\star}$}, Zhao Xu, Weihua Luo\\ \bf Alibaba International Digital Commerce \\
\vspace{1mm} $^{\star}$ Corresponding Author: Longyue Wang\\
\vspace{1mm} $^{\dagger}$ Equal Contribution\\
}

\begin{abstract}
{\bf \large Abstract}\vspace{1mm}

Deep research agents autonomously conduct open-ended investigations, integrating complex information retrieval with multi-step reasoning across diverse sources to solve real-world problems.
To sustain this capability on long-horizon tasks, reliable verification is critical during both training and inference. A major bottleneck in existing paradigms stems from the lack of explicit verification mechanisms in QA data synthesis, trajectory construction, and test-time scaling. Errors introduced at each stage propagate downstream and degrade the overall agent performance. To address this, we present Marco DeepResearch, a deep research agent optimized with a verification-centric framework design at three levels: \textbf{(1)~QA Data Synthesis:} We introduce verification mechanisms to graph-based and agent-based QA synthesis to control question difficulty while ensuring answers are unique and correct; \textbf{(2)~Trajectory Construction:} We design a verification-driven trajectory synthesis method that injects explicit verification patterns into training trajectories; and \textbf{(3)~Test-time scaling:} We use Marco DeepResearch itself as a verifier at inference time and effectively improve performance on challenging questions. Extensive experimental results demonstrate that our proposed Marco DeepResearch agent significantly outperforms 8B-scale deep research agents on most challenging benchmarks, such as BrowseComp and BrowseComp-ZH. Crucially, under a maximum budget of 600 tool calls, Marco DeepResearch even surpasses or approaches several 30B-scale agents, like Tongyi DeepResearch-30B.
\\
\\
\makebox[1pt][l]{\parbox{\textwidth}{\raggedright\begin{tabular}{@{} l l @{}} \raisebox{-0.5em}{\includegraphics[height=1.6em]{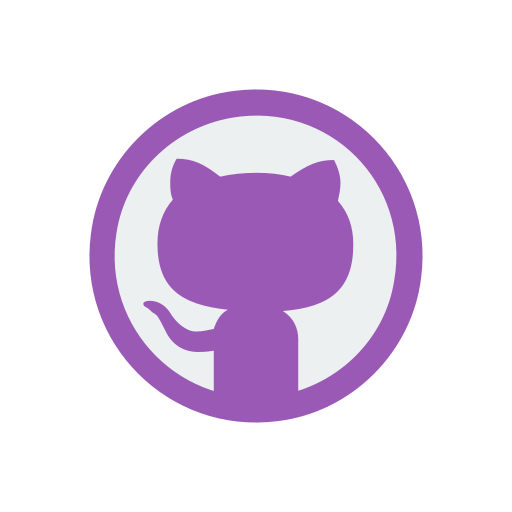}} & \small\url{https://github.com/AIDC-AI/Marco-DeepResearch}
\end{tabular}}}

\end{abstract}

\begin{document}

\maketitle


\section{Introduction}
\label{sec:introduction}

\begin{figure}[ht]
    \centering
    \includegraphics[width=\textwidth]{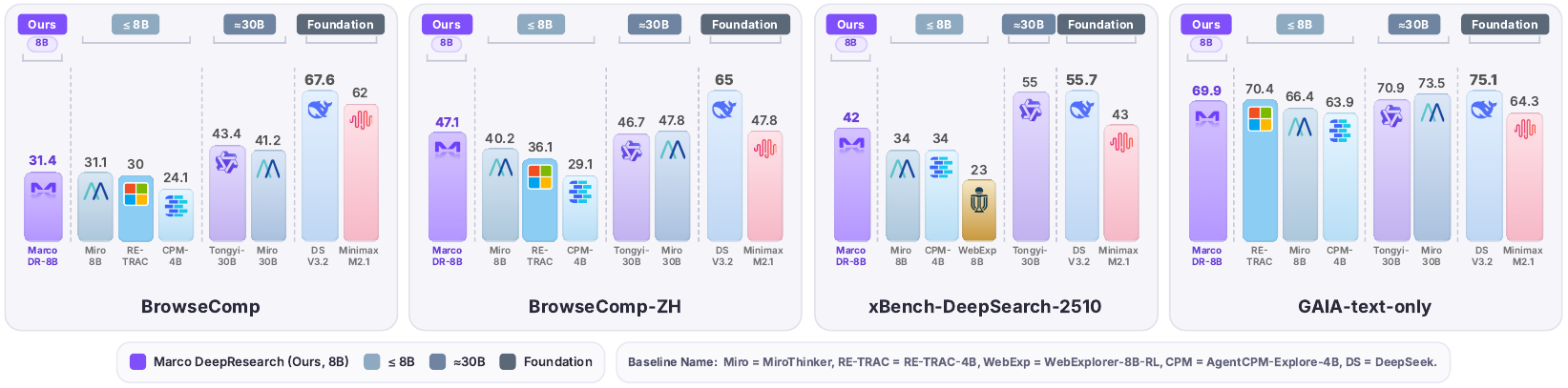}
    \caption{Benchmark performance of our proposed Marco DeepResearch 8B-scale agent.}
    \label{fig:benchmark}
\end{figure}

Large language models (LLMs) have enabled tool-augmented agents that can autonomously reason and interact with external environments~\cite{team2025tongyi,huang2025deep}. Within this line, Deep research agents~\cite{OpenAI-DeepResearch, Gemini-DeepResearch} have attracted broad attention: proprietary systems such as OpenAI Deep Research~\cite{OpenAI-DeepResearch} and Gemini Deep Research~\cite{Gemini-DeepResearch} demonstrate exceptional information-seeking capabilities for solving complex tasks in real-world scenarios, while open-source systems like MiroThinker~\cite{miromind2025mirothinker}, Tongyi DeepResearch~\cite{team2025tongyi}, and AgentCPM-Explore~\cite{AgentCPMExplore2026} have rapidly narrowed the gap and in some settings matched or surpassed proprietary alternatives~\cite{mialon2023gaiabenchmarkgeneralai,wei2025browsecompsimplechallengingbenchmark,phan2025humanitysexam}. This progress is largely driven by advances in data synthesis~\cite{li2025websailornavigatingsuperhumanreasoning,tao2025webshaperagenticallydatasynthesizing}, reinforcement learning~\cite{shao2024deepseekmathpushinglimitsmathematical,dong2025agenticreinforcedpolicyoptimization}, and test-time scaling~\cite{xu2026wideseekr1exploringwidthscaling,li2025chainofagentsendtoendagentfoundation}.

Despite this progress, current deep research agents face a critical bottleneck: the lack of explicit verification across the following three essential stages, leading to error propagation and overall performance degradation of agents~\cite{c3b1c1524d8941978df8bd55d513abc7,lan2024traininglanguagemodelscritique,lan2024criticevalevaluatinglargelanguage}:
(1) \textbf{QA Data Synthesis}: most works synthesize QA samples from graph-based~\cite{wu2025webwalkerbenchmarkingllmsweb} or agent-based web exploration~\cite{xu2026sagesteerableagenticdata}, but entity obfuscation—the most widely adopted technique in existing pipelines—often yields non-unique or incorrect answers~\cite{meituanlongcatteam2026longcatflashthinking2601technicalreport}, undermining supervision quality and propagating errors to downstream trajectory construction; 
(2) \textbf{Trajectory construction}: most existing works rely on strong teacher models  to generate ReAct-style trajectories that can directly reach correct final answers, but these trajectories usually lack explicit verification~\cite{yao2023reactsynergizingreasoningacting}; as a result, trained agents tend to accept early low-quality results, under-explore high-value alternatives~\cite{hu2025webcotenhancingwebagent,wan2026inference}; 
and (3) \textbf{Test-time scaling}: systems generally lack explicit verification for both intermediate steps and the final answers during inference; consequently, flawed intermediate states and incorrect conclusions propagate unchecked, leading agents to accept early errors rather than triggering verifier-guided behaviors to effectively scale test-time compute~\cite{wan2026inference}.

To address these verification gaps, we present \textbf{Marco DeepResearch}, an efficient 8B-scale deep research agent with three improvements across these stages: (1) \textbf{Verified Data Synthesis} (Section~\ref{sec:verified-data}): where we introduce an explicit verification mechanisms to graph-based and agent-based QA synthesis methods so that question-answer pairs are carefully checked, ensuring their difficulties, uniqueness and correctness; (2) \textbf{Verification-Driven Trajectory Construction} (Section~\ref{sec:verified-framework}): where we introduce a specialized verifier agent to verify the answers of sub-tasks and final answer using web search tools, providing more explicit verification patterns into single-agent and multi-agent trajectories; and (3) \textbf{Verifier-Guided Test-Time Scaling} (Section~\ref{sec:verified-inference}): we use the Marco DeepResearch agent itself as a verifier and continue reasoning on challenging questions under a controlled compute budget, thereby more effectively unlocking the potential of test-time scaling.


With these optimizations, we synthesize high-quality trajectories and train the Marco DeepResearch agent based on the Qwen3-8B base model~\cite{yang2025qwen3technicalreport}, and evaluate it on six  deep search benchmarks like BrowseComp~\cite{wei2025browsecompsimplechallengingbenchmark}, BrowseComp-ZH~\cite{zhou2025browsecompzhbenchmarkingwebbrowsing}, and GAIA~\cite{mialon2023gaiabenchmarkgeneralai}.
Specifically, Marco DeepResearch outperforms 8B-scale deep research agents on most challenging deep benchmarks such as BrowseComp~\cite{miromind2025mirothinker,AgentCPMExplore2026}. Moreover, under a budget of up to 600 tool calls~\cite{miromind2025mirothinker}, our proposed Marco DeepResearch agent surpasses MiroThinker-v1.0-8B on BrowseComp-ZH and matches or exceeds several 30B-scale agents, including Tongyi DeepResearch-30B~\cite{team2025tongyi} and MiroThinker-v1.0-30B~\cite{miromind2025mirothinker}. Ablation studies further prove the contributions of our designs for optimizing Marco DeepResearch.


\section{Related Work}
\label{sec:related}

\paragraph{Deep Research agent systems.}

LLM-based agent systems have demonstrated profound potential and versatility across a wide spectrum of complex tasks~\cite{yang2025hscodecomprealisticexpertlevelbenchmark,lan2025deepwidesearchbenchmarkingdepthwidth,ye2026umemunifiedmemoryextraction,team2025tongyi,5team2026glm5vibecodingagentic}. Building on this foundation, \textit{Deep Research} has emerged as a frontier application, with commercial systems~\cite{OpenAI-DeepResearch,Gemini-DeepResearch} showcasing remarkable capabilities in conducting open-ended investigations and synthesizing comprehensive reports. At the very core of these sophisticated research systems lies \textit{Deep Search} (i.e., agentic information seeking)---the indispensable engine that enables agents to autonomously plan, navigate multi-turn web interactions, and extract reasoning-driven evidence~\cite{lan2025deepwidesearchbenchmarkingdepthwidth,huang2025deep,lan2026tableassearchformulatelonghorizonagentic,wong2025widesearchbenchmarkingagenticbroad}. However, despite rapid advancements and the emergence of open-source deep research agents~\cite{miromind2025mirothinker,team2025tongyi}, critical bottlenecks persist in data quality and inference-time strategies during long horizons.


\paragraph{Data synthesis for deep research agents.}
High-quality synthetic data is the key to the agentic search capabilities~\cite{team2025tongyi,hu2025stepdeepresearchtechnicalreport,tao2025webshaperagenticallydatasynthesizing}. Current approaches to agentic data synthesis mainly follow two paradigms: (1) graph-based methods traverse knowledge graphs to synthetic multi-hop QA data~\cite{miromind2025mirothinker}; (2) agent-based methods use agents explore real web environments~\cite{xu2026sagesteerableagenticdata} for data synthesis. Despite their differences, both paradigms face a common and fundamental challenge: automatically synthesizing difficult QA pairs with unique and correct answers~\cite{meituanlongcatteam2026longcatflashthinking2601technicalreport}. To address this issue, we design a verification-driven method to improve QA data quality. 

\paragraph{Trajectory construction.}
The ReAct paradigm~\cite{yao2023reactsynergizingreasoningacting} serves as the foundation of most current agentic systems.
Recent works have improved upon ReAct through procedural planning~\cite{wang2023planandsolvepromptingimprovingzeroshot}, multi-agent orchestration~\cite{wong2025widesearchbenchmarkingagenticbroad,lan2026tableassearchformulatelonghorizonagentic}, and context management~\cite{deepagent}.
However, these frameworks share a critical limitation: the absence of \emph{explicit verification} during interactions~\cite{wan2026inference}.
In long-horizon information seeking, agents must navigate massive search spaces where intermediate results are often noisy or misleading.
Without a dedicated verification mechanism, agents are prone to accepting the first plausible-looking answer and terminating exploration prematurely, even when the result is incorrect~\cite{wan2026inference}.
To address this limitation, we introduce an explicit verification mechanisms for both intermediate search results and final answers, designed to effectively teach the model robust verification behaviors.

\paragraph{Test-time scaling.}
At test time, deep research agents solve complex problems through extensive interactive exploration of web environments.
Effective test-time scaling strategies can significantly enhance agent performance by allocating more computation at inference~\cite{snell2024scalingllmtesttimecompute,miromind2025mirothinker,zhu2026retracrecursivetrajectorycompression,miromindteam2026mirothinker17h1heavyduty}.
While current test-time scaling approaches for agentic search primarily focus on multi-agent coordination~\cite{lan2026tableassearchformulatelonghorizonagentic} and context summarization~\cite{wu2025resumunlockinglonghorizonsearch,zhu2026retracrecursivetrajectorycompression}, the role of explicit verification as a systematic test-time scaling strategy for trained deep search agents remains largely unexplored~\cite{du2026openseekerdemocratizingfrontiersearch,wan2026inferencetimescalingverificationselfevolving}.
We address this gap by using Marco DeepResearch itself as a verifier at inference time, realizing effective test-time scaling by extending reasoning turns.


\section{Verified Data Synthesis}
\label{sec:verified-data}

In this section, we apply explicit verification to QA data synthesis to ensure the quality and answer uniqueness while keeping their difficulties.
High-quality QA data is essential for both trajectory synthesis and optimization.
A common bottleneck in existing approaches is \textit{answer non-uniqueness}: to increase question difficulty, most methods obfuscate entity information in multi-hop questions~\cite{miromind2025mirothinker, xu2026sagesteerableagenticdata}, which inevitably introduces ambiguity and may result in low-quality questions with multiple valid answers.
When such data is used as ground truth, the training becomes biased and unstable.
We address this problem through two complementary synthesis pipelines: graph-based and agent-based, each incorporating explicit verification to guarantee answer quality.

\begin{figure*}[t]
    \includegraphics[width=\textwidth]{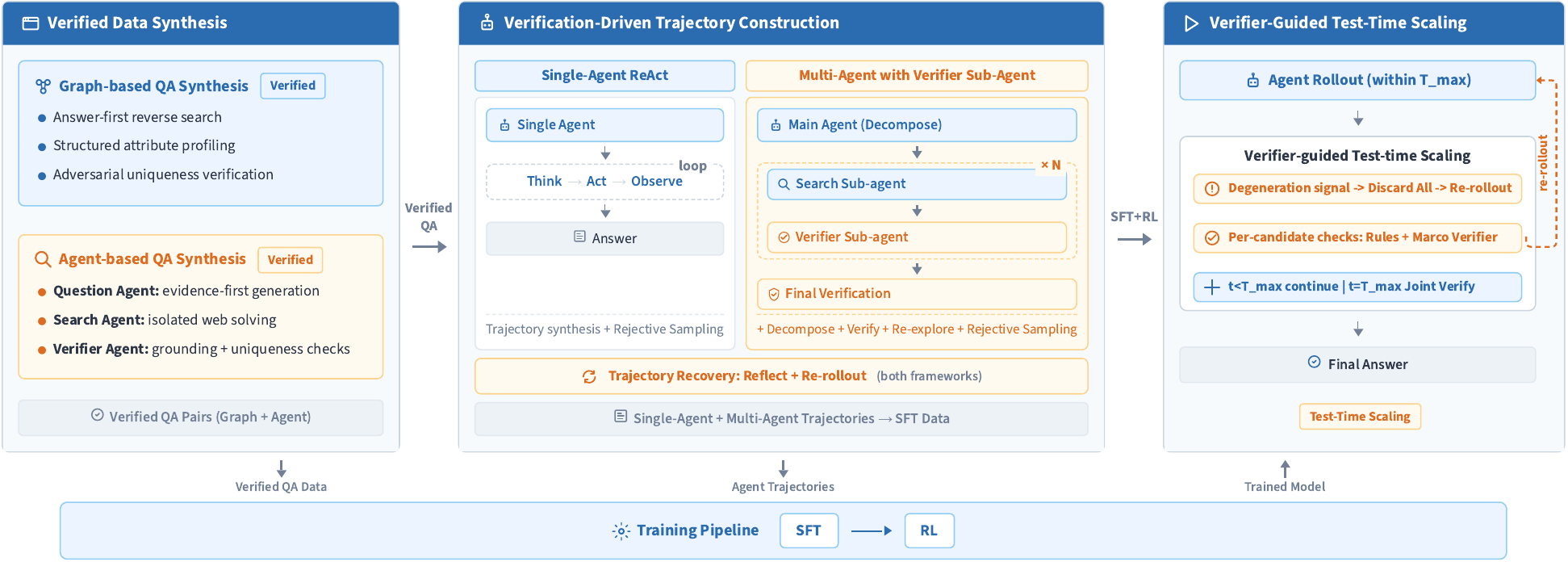}
    \caption{Overall framework of our verification-center design for Marco DeepResearch agents.}
    \label{fig:overview}
\end{figure*}


\subsection{Graph-Based Synthesis with Adversarial Verification}
\label{sec:qa-graph}

Most graph-based QA synthesis methods still face a core difficulty: it is hard to jointly guarantee search depth, answer uniqueness, correctness, and entity leakage control in QA construction~\cite{meituanlongcatteam2026longcatflashthinking2601technicalreport}. To address this, we introduce a unified paradigm of \textit{answer-first reverse construction with adversarial verification} in graph-based QA synthesis, organized as an iterative loop:

\paragraph{Answer entity sampling.}
We first sample answer entities under structural and content constraints on knowledge-graph (e.g., moderate connectivity, sufficient document evidence, and valid predecessor nodes), ensuring tasks necessitate multi-hop reasoning while avoiding trivial common-knowledge shortcuts.

\paragraph{Structured attribute profiling.}
Given documents linked to the each sampled answer entity, we leverage frontier models to extract a structured attribute profile over five dimensions: spatial, temporal, numerical, categorical, and entity-relation features. This profile provides the candidate constraints for controlled obfuscation and question construction.

\paragraph{Reverse path search.}
Starting from the selected answer entity, we search backward for intermediate evidence nodes using complementary graph-structure search and content-matching search (attribute keywords matching). Then, strong LLMs are used to select a small set of high-quality, diverse intermediates (4 to 8 intermediates) to form a robust multi-hop reasoning chain.

\paragraph{Adversarial answer uniqueness verification.}
After getting a searched path containing the answer entity, we apply an adversarial verification process to ensure answer uniqueness. This is the key verification process to ensure answer uniqueness and difficulty of synthesized QA pairs, which is an iterative three-role process with a \textit{Generator}, an \textit{Attacker}, and an \textit{Analyzer}. 
The Generator first initializes 2--3 obfuscated constraints from the attribute profile; the Attacker then searches for counterexample entities that satisfy all current constraints but are not the target answer. If no counterexample is found and the constraint count is above a minimum threshold, the loop converges; otherwise, the Analyzer adds new discriminative constraints and returns control to the Attacker.
This loop runs for at most 10 rounds. Its convergence follows a monotonicity principle: each round appends at least one new constraint, and each added constraint removes at least part of the counterexample set. As a result, the final constraint set provides high-confidence answer uniqueness for the target entity.

Finally, after convergence, we convert constraints into natural-language multi-hop questions and apply leakage checks to obscure key entities. Samples exhibiting leakage, or those solvable by frontier models without search and consistency checks, are excluded from our training set.

\subsection{Agent-Based Web Exploration Synthesis}
\label{sec:qa-agent}

Compared with graph-based methods, agent-based QA synthesis significantly enhances data realism and broadens domain coverage~\cite{miromind2025mirothinker,xu2026sagesteerableagenticdata}. Motivated by these advantages, we also construct QA data via agent-based web exploration, empowering agents to autonomously navigate real-world web environments to formulate real-world, complex, multi-hop questions~\cite{xu2026sagesteerableagenticdata, tao2025webshaperagenticallydatasynthesizing}. Despite its benefits, this dynamic setting inevitably introduces common failure cases, such as factual hallucinations, ambiguous answers, and pseudo multi-hop questions that are easily bypassed by shortcut retrieval~\cite{xu2026sagesteerableagenticdata}.
To control these failures, we design a Generation--Execution--Verification loop with a question agent, a search agent, and a verification agent. The key design is to separate question construction from independent solving and then enforce strict third-party verification before data acceptance.

\paragraph{Evidence-first question construction.}
Instead of forward generation, question agent first explores the open web to build an evidence graph, and constructs questions from verified evidence. During construction, it applies entity obfuscation and diverse reasoning topologies (e.g., convergent and conjunctive constraints) to reduce one-step shortcut matching while controlling target difficulty~\cite{xu2026sagesteerableagenticdata}.

\paragraph{Multi-stage quality verification.}
We employ a multi-stage filtering pipeline: a verification agent ensures factual consistency and evidence grounding, while a closed-book filter excludes questions solvable without retrieval. Remaining candidates are solved by an independent search agent, with final verification confirming reasoning depth aligns with target difficulty and no alternative valid answers satisfy the constraints~\cite{xu2026sagesteerableagenticdata}.


\paragraph{Diagnosis Iterative Optimization.}
When a sample fails at any stage, we do not simply discard it. Instead, verification agent provides structured diagnostic feedback (e.g., under-constrained question, shortcut path, insufficient depth, or evidence conflict), and question agent performs targeted updates on evidence selection, constraint design, and question structure. This diagnosis--revision loop continues until the sample jointly satisfies groundedness, uniqueness, and empirical difficulty requirements, improving data efficiency while maintaining strict quality control.

We combine the above two pipelines with additional synthesis strategies to maximize data diversity across problem types, domains, and difficulty levels. To validate data quality, we manually reviewed 100 samples. Fewer than 10\% had clear question-answer mismatch, while the remaining QA samples are valid but challenging. This result shows that our proposed methods could generate high-quality and challenging dataset for optimizing agents.


\section{Verification-Driven Trajectory Construction}
\label{sec:verified-framework}

Single-agent ReAct is still the dominant trajectory synthesis recipe in current deep research systems~\cite{li2025websailornavigatingsuperhumanreasoning,miromind2025mirothinker}. However, this pipeline typically does not explicitly verify key intermediate results, so errors made in early steps can directly propagate and accumulate, degrading final performance. We therefore argue that high-quality trajectories should contain explicit verification patterns, including both intermediate checks for sub-task outputs and final checks for the proposed answer. Prior work~\cite{wan2026inferencetimescalingverificationselfevolving} on deep-search benchmarks also suggests that, for needle-in-a-haystack tasks, direct solving is difficult while answer verification conditioned on the question (or sub-question) is relatively reliable~\cite{mialon2023gaiabenchmarkgeneralai,wei2025browsecompsimplechallengingbenchmark}.
To capitalize on this easy-to-verify property, we introduce two complementary designs for trajectory construction: multi-agent verified synthesis and verification-reflection re-rollout.

\paragraph{Multi-agent with Verification.}
As shown in Figure~\ref{fig:overview}, we design a three-role framework with a main agent, a search sub-agent, and a verifier sub-agent. The main agent decomposes a complex problem into sub-tasks and aggregates sub-results into a final answer. The search sub-agent solves each sub-task. The verifier agent then performs independent third-party validation with web tools for both sub-task outputs and the final proposed answer. If verification fails, the corresponding step is revised and re-executed, so trajectories explicitly record verification-driven correction behavior.
Finally, the multi-agent trajectories are converted into a single-agent ReAct-style trajectories for training~\cite{li2025chainofagentsendtoendagentfoundation}.

\paragraph{Verification-Reflection Re-rollout on Failed Trajectories.}
We also collect trajectories with incorrect final answers and invoke a verifier agent to diagnose failure causes and produce actionable feedback~\cite{zhu2026retracrecursivetrajectorycompression}. Conditioned on this feedback, we re-rollout the failed trajectories and keep trajectories that are recovered to correct answers.


\section{Verifier-Guided Test-Time Scaling}
\label{sec:verified-inference}

Current test-time scaling for deep research agents mainly increases interaction rounds or rollout budget~\cite{miromind2025mirothinker}. While this can improve coverage, blindly scaling turns often accumulates early tool errors and noisy intermediate conclusions, which reduces reliability on long-horizon search tasks~\cite{miromindteam2026mirothinker17h1heavyduty}. Therefore, we propose \textbf{Verifier-Guided Test-time Scaling} that adds explicit verification into inference-time scaling and uses Marco DeepResearch itself as a verifier. By combining the \textit{Discard All} context management strategy with verification~\cite{deepseekai2025deepseekv32pushingfrontieropen}, we realize more effective test-time scaling under a fixed maximum interaction budget $T_{\max}$.

\paragraph{Discard All.}
During a rollout, once predefined degeneration signals are triggered (e.g., reaching max steps or failing to solve questions), we apply \textit{Discard All} context management strategy: remove accumulated tool-call history and intermediate reasoning outputs, keep only the original query and the system prompt, and restart from a fresh context. This reset mechanism allows the agent to explore new search paths and reducing error propagation along a single trajectory~\cite{deepseekai2025deepseekv32pushingfrontieropen}.

\paragraph{Verifier-Guided Test-time Scaling.}
Whenever the agent produces a candidate answer, we conduct rule-based checks and agent-as-a-judge using Marco DeepResearch~\cite{wan2026inferencetimescalingverificationselfevolving,zhuge2024agentasajudgeevaluateagentsagents}. If $t<T_{\max}$~\cite{miromind2025mirothinker,chen2026searchmorethinkless}, the agent can continue exploring and propose additional candidates; each candidate is verified independently. When $t=T_{\max}$ or the process reaches a convergence condition, we perform \textit{Joint Verify} over all candidates and generate the final answer for the question.

These two components are complementary: \textit{Discard All} improves trajectory quality by resetting degraded contexts, while \textit{Verifier-guided Test-time Scaling} improves answer quality. Together, they realize more effective test-time scaling without changing model parameters, and unlock stronger inference-time gains on hard questions.


\section{Training Pipeline}
\label{sec:training}

The training pipeline consists of Supervised Fine-Tuning and Reinforcement Learning.

\subsection{Supervised Fine-Tuning}
\label{sec:training-sft}

\paragraph{Training objective.}
We train with token-level cross-entropy and apply a loss mask so that only assistant response tokens contribute to optimization $\mathcal{L}_{\text{SFT}}(\theta) = - \sum_{t=1}^{T} m_t \log P_\theta(x_t \mid x_{<t})$, where the mask is defined as
\begin{equation}
m_t=
\begin{cases}
1, & t \in \mathcal{T}_{\text{assistant}},\\
0, & t \in \mathcal{T}_{\text{instruction}} \cup \mathcal{T}_{\text{tool\_response}}.
\end{cases}
\end{equation}
That is, instruction and tool response content are masked out.

\subsection{Reinforcement Learning}
\label{sec:training-rl}

Starting from the SFT checkpoint, we optimize the policy with Group Relative Policy Optimization (GRPO)~\cite{shao2024deepseekmathpushinglimitsmathematical}, where updates are driven by within-group relative advantages. Concretely, for each query $q$, we sample a group of $G$ rollouts $\{o_i\}_{i=1}^{G}$ from the old policy $\pi_{\theta_{\text{old}}}$ and optimize
\begin{equation}
\begin{aligned}
\mathcal{J}_{\text{GRPO}}(\theta) = \mathbb{E}_{\substack{q \sim P(Q) \\ \{o_i\}_{i=1}^{G} \sim \pi_{\theta_{\text{old}}}}} \Bigg[ & \frac{1}{G}\sum_{i=1}^{G} \min \Big( r_i(\theta)\hat{A}_i,\, \mathrm{clip}\!\left(r_i(\theta), 1-\epsilon, 1+\epsilon\right)\hat{A}_i \Big) \\
& - \beta\, \mathbb{D}_{\text{KL}}\!\left[\pi_{\theta} \,\|\, \pi_{\text{ref}}\right] \Bigg]
\end{aligned}
\end{equation}
where $r_i(\theta) = \frac{\pi_{\theta}(o_i|q)}{\pi_{\theta_{\text{old}}}(o_i|q)}$ denotes the importance sampling ratio. The relative advantage is computed by reward normalization within each group:
\begin{equation}
\hat{A}_i = \frac{r_i - \mathrm{mean}\left(\{r_j\}_{j=1}^{G}\right)}{\mathrm{std}\left(\{r_j\}_{j=1}^{G}\right)}.
\end{equation}

We adopt an outcome-based reward, and balance reward quality and computational cost by using a two-stage LLM-as-Judge pipeline: a fast primary judge (Qwen-Turbo-Latest) evaluates all samples, and uncertain or low-confidence cases are escalated to a secondary judge (GPT-4.1) for re-evaluation:
\begin{equation}
r(q,o)=
\begin{cases}
1, & \text{if }\mathcal{J}(o,a^{*})=\text{correct},\\
0, & \text{otherwise},
\end{cases}
\end{equation}
where $q$ is the input query, $o$ is the generated output, $a^{*}$ is the reference answer, and $\mathcal{J}(\cdot)$ is the judging function.


\section{Experimental Setup}
\label{sec:exp-setup}

\paragraph{Benchmarks.}
We evaluate our proposed Marco DeepResearch agent on six deep search benchmarks: (1) \textbf{BrowseComp}~\cite{wei2025browsecompsimplechallengingbenchmark}:
Measuring agent's information seeking capability by navigating the web; (2) \textbf{BrowseComp-ZH}~\cite{zhou2025browsecompzhbenchmarkingwebbrowsing}: Chinese counterpart evaluating agentic information seeking; (3) \textbf{GAIA (text-only)}~\cite{mialon2023gaiabenchmarkgeneralai}: Real-world multi-step questions for general AI assistants; (4) \textbf{xBench-DeepSearch}~\cite{xBench-DeepSearch}: Deep search across diverse domains\footnote{Both 2505 and 2510 splits are evaluated.}; (5) \textbf{WebWalkerQA}~\cite{wu2025webwalkerbenchmarkingllmsweb}: Multi-step web navigation and information extraction; and (6) \textbf{DeepSearchQA}~\cite{gupta2026deepsearchqabridgingcomprehensivenessgap}: Evaluating exhaustive answer set generation through multi-source retrieval, entity resolution, and stopping criteria reasoning.

\paragraph{Baselines.}
We compare against three groups of state-of-the-art baselines: (1) \textbf{Foundation models with tools}: GLM-4.7~\cite{5team2026glm5vibecodingagentic}, Minimax-M2.1, DeepSeek-V3.2~\cite{deepseekai2025deepseekv32pushingfrontieropen}, Kimi-K2.5~\cite{kimiteam2026kimik25visualagentic}, Claude-Sonnet/Opus, OpenAI-o3, GPT-5 High and Gemini-3-Pro;  (2) \textbf{Trained agents $\geq$ 30B-scale}: Tongyi DeepResearch~\cite{team2025tongyi}, WebSailor-v2~\cite{li2025websailornavigatingsuperhumanreasoning}, MiroThinker-v1.0/v1.5/v1.7~\cite{miromind2025mirothinker,miromindteam2026mirothinker17h1heavyduty}, DeepMiner~\cite{tang2025turnlimitstrainingdeep}, OpenSeeker-30B-SFT~\cite{du2026openseekerdemocratizingfrontiersearch}, and SMTL~\cite{chen2026searchmorethinkless}; and (3) \textbf{Trained agents $\leq$ 8B-scale}: MiroThinker-v1.0-8B, WebExplorer-8B-RL~\cite{liu2025webexplorerexploreevolvetraining}, AgentCPM-Explore-4B~\cite{AgentCPMExplore2026} and RE-TRAC-4B~\cite{zhu2026retracrecursivetrajectorycompression}.

\paragraph{Training Data.}
Our training corpus contains two sources: \textbf{(1) Open-source data}, including 2WikiMultihopQA~\cite{ho2020constructingmultihopqadataset}, BeerQA~\cite{qi2021answeringopendomainquestionsvarying}, ASearcher~\cite{gao2025turnsunlockinglonghorizonagentic}, DeepDive~\cite{lu2025deepdiveadvancingdeepsearch}, QA-Expert-Multi-Hop-QA~\cite{qa-expert}, and REDSearcher~\cite{chu2026redsearcherscalablecostefficientframework}; and \textbf{(2) Synthetic data}, including (i) real-world e-commerce business-development datasets from our inner applications~\cite{lan2025deepwidesearchbenchmarkingdepthwidth,yang2025hscodecomprealisticexpertlevelbenchmark,lan2026tableassearchformulatelonghorizonagentic} and (ii) data synthesized by our verified data synthes. In details, we collect over 12K synthesized graph-based QA and agent-based QA samples. Moreover, we hold-out over 2K high-quality QA samples for RL training. Trajectory data are synthesized using frontier foundation models, including Qwen3.5-Plus, GLM-5, and Kimi-K2, among others, followed by data cleaning (e.g., tool-call error correction).

\paragraph{Implementation Details.}
We use Qwen3-8B as the backbone. YaRN is used to extend the context window to 128K. Supervised fine-tuning and RL are conducted on 64 A100 GPUs using Megatron~\cite{megatron-lm}. 
To improve system efficiency and stability, we use Redis-based caching for repeated queries/pages, exponential-backoff retries for transient failures, asynchronous non-blocking tool calls, asynchronous reward computation pipelined with model updates, and synchronized deployment of the WebVisit summary model as an independent training-cluster service~\cite{AgentCPMExplore2026}.
As for the evaluation details, we follow previous works~\cite{miromind2025mirothinker} and evaluate Marco DeepResearch agent under a maximum budget of 600 tool calls. Decoding uses temperature $0.7$, top-$p$ $0.95$, and a maximum generation length of 16{,}384 tokens.

\section{Experimental Results}
\subsection{Main Results}
\label{sec:exp-main}
\begin{table}[t]
\centering
\caption{Performance on deep search benchmarks. Best open-source results are \textbf{bolded} and the second-best are \underline{underlined}. Results marked with $^\star$ are evaluated using our implementation.}
\label{tab:main-results}
\small
\setlength{\tabcolsep}{4pt}
\begin{adjustbox}{max width=\textwidth}
\begin{tabular}{l c c c c c c c}
\toprule
\textbf{Model} & \begin{tabular}{@{}c@{}}\textbf{Browse} \\ \textbf{Comp}\end{tabular} & \begin{tabular}{@{}c@{}}\textbf{Brows} \\ \textbf{Comp-ZH}\end{tabular} & \begin{tabular}{@{}c@{}}\textbf{GAIA} \\ \textbf{text-only}\end{tabular} & \begin{tabular}{@{}c@{}}\textbf{Web} \\ \textbf{WalkerQA}\end{tabular} & \begin{tabular}{@{}c@{}}\textbf{xBench-} \\ \textbf{DS-2505}\end{tabular} & \begin{tabular}{@{}c@{}}\textbf{xBench-} \\ \textbf{DS-2510}\end{tabular} & \begin{tabular}{@{}c@{}}\textbf{Deep} \\ \textbf{SearchQA}\end{tabular} \\
\midrule
\multicolumn{6}{l}{\textbf{Foundation Models with Tools}} \\
\midrule
GLM-4.7 & 67.5 & 66.6 & 61.9 & -- & 72.0 & 52.3 & -- \\
Minimax-M2.1 & 62.0 & 47.8 & 64.3 & -- & 68.7 & 43.0 & --\\
DeepSeek-V3.2 & 67.6 & 65.0 & 75.1 & -- & 78.0 & 55.7 & 60.9 \\
Kimi-K2.5 & 74.9 & 62.3 & -- & -- & -- & 46.0 & 77.1 \\
Claude-4-Sonnet & 12.2 & 29.1 & 68.3 & 61.7 & 64.6 & -- & -- \\
Claude-4.5-Opus & 67.8 & 62.4 & -- & -- & -- & -- & 80.0 \\
OpenAI-o3 & 49.7 & 58.1 & -- & 71.7 & 67.0 & -- & -- \\
OpenAI GPT-5 High & 54.9 & 65.0 & 76.4 & -- & 77.8 & 75.0 & 79.0 \\
Gemini-3.0-Pro & 59.2 & 66.8 & -- & -- & -- & 53.0 & 76.9 \\
\midrule
\multicolumn{6}{l}{\textbf{Trained Agents ($\geq$30B)}} \\
\midrule
MiroThinker-v1.7-mini & 67.9 & 72.3 & 80.3 & -- & -- & 57.2 & 67.9 \\
MiroThinker-v1.5-235B & 69.8 & 71.5 & 80.8 & -- & 77.1 & -- & --\\
MiroThinker-v1.5-30B & 56.1 & 66.8 & 72.0 & -- & 73.1& -- & --\\
MiroThinker-v1.0-72B & 47.1 & 55.6 & 81.9 & 62.1 & 77.8 & -- & -- \\
MiroThinker-v1.0-30B & 41.2 & 47.8 & 73.5 & 61.0 & 70.6 & -- & -- \\
SMTL-30B-300 & 48.6 & -- & 75.7 & 76.5 & 82.0 & -- & --\\
Tongyi-DR-30B & 43.4 & 46.7 & 70.9 & 72.2 & 75.0 & 55.0 & --\\
WebSailor-V2-30B & 35.3 & 44.1 & 74.1 & -- & 73.7& -- & -- \\
DeepMiner-32B-RL & 33.5 & 40.1 & 58.7 & -- & 62.0 & -- & --\\
OpenSeeker-30B-SFT & 29.5 & 48.4 & -- & -- & 74.0 & -- & --\\
\midrule
\multicolumn{6}{l}{\textbf{Trained Agents ($\leq$8B)}} \\
\midrule
AgentCPM-Explore-4B & 24.1 & 29.1 & 63.9 & \underline{68.1} & \underline{70.0} & \underline{34.0}$^{\star}$ & \underline{32.8}$^{\star}$\\
WebExplorer-8B-RL & 15.7 & 32.0 & 50.0 & 62.7 & 53.7 & 23.0$^{\star}$ & 17.8$^{\star}$\\
RE-TRAC-4B & 30.0 & 36.1 & \textbf{70.4} & -- & 74.0 & -- & --\\
MiroThinker-v1.0-8B & \underline{31.1} & \underline{40.2} & 66.4 & 60.6 & 60.6 & \underline{34.0}$^{\star}$ & \textbf{36.7}$^{\star}$ \\
\rowcolor{blue!10} \textbf{Marco-DR-8B (Ours)} & \textbf{31.4} & \textbf{47.1} & \underline{69.9} & \textbf{69.6} & \textbf{82.0} & \textbf{42.0} & 29.2 \\
\bottomrule
\end{tabular}
\end{adjustbox}
\end{table}

Table \ref{tab:main-results} demonstrates that Marco-DeepResearch-8B outperforms other 8B-scale open-source deep search trained agents on most of benchmarks. 
Specifically, it attains the highest scores in its size category on exploration-heavy tasks, including BrowseComp (31.4), BrowseComp-ZH (47.1), WebWalkerQA (69.6), and xBench-DeepSearch (82.0 on the 2505 split and 42.0 on the 2510 split). For the remaining three benchmarks, our proposed Marco-DeepResearch agent remains highly competitive, missing the top score on GAIA text-only by a marginal 0.5 points compared to RE-TRAC-4B.
Notably, Marco-DeepResearch-8B approaches and even surpasses several competitive 30B-scale deep search agents on several benchmark, like Tongyi-DeepResearch. These results validate the efficacy of our proposed QA data synthesis, trajectory construction methods and test-time scaling strategy, proving that our optimized 8B model can effectively close the performance gap with massive foundation models in complex web navigation and information-seeking tasks.

\subsection{Analysis}
\label{sec:exp-analysis}

We conduct detailed analysis from five aspects: (1) data statistics analysis; (2) effect of QA data verification; (3) ablation study on verification-driven trajectory construction; (4) improvement of reinforcement learning; (5) ablation study on verifier-guided test-time scaling; and (6) context window extension during training.

\paragraph{Data statistics analysis.}

\begin{figure}[h]
    \centering
    \begin{minipage}[t]{0.49\linewidth}
        \centering
        \includegraphics[width=\linewidth]{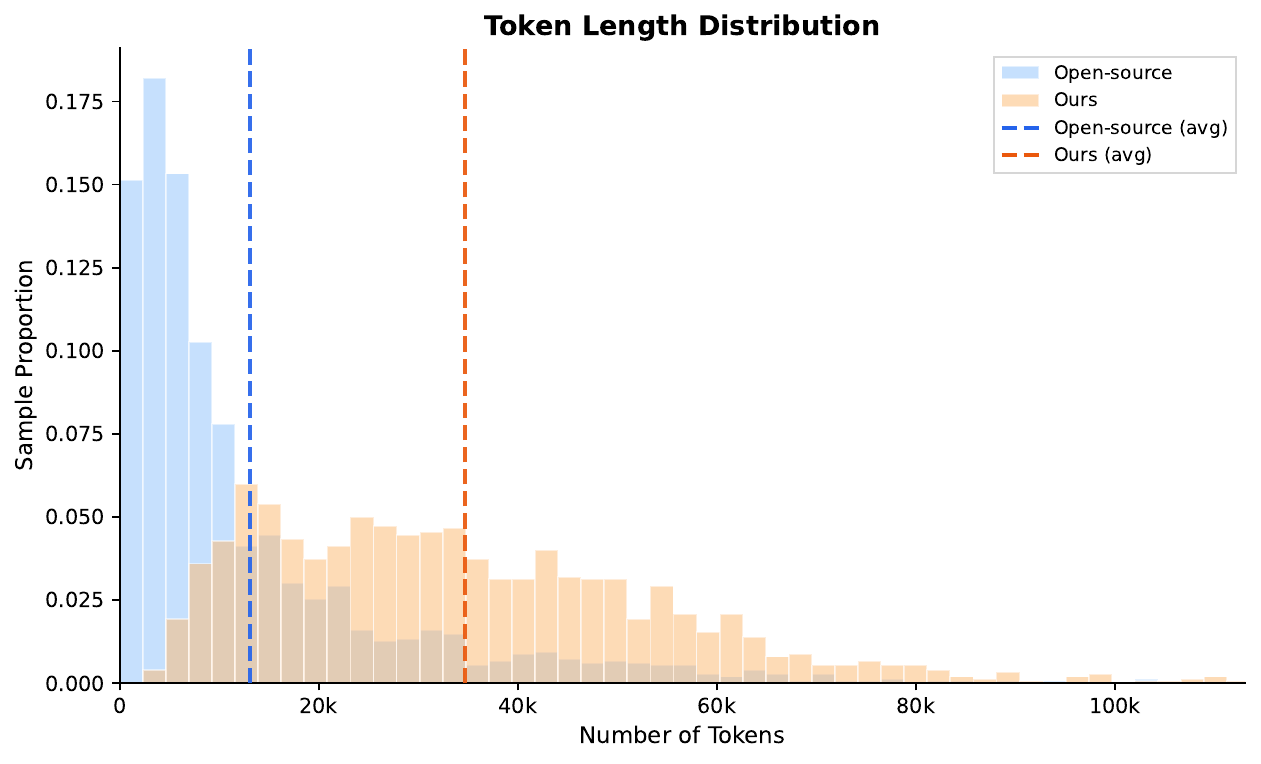}
    \end{minipage}
    \hfill
    \begin{minipage}[t]{0.49\linewidth}
        \centering
        \includegraphics[width=\linewidth]{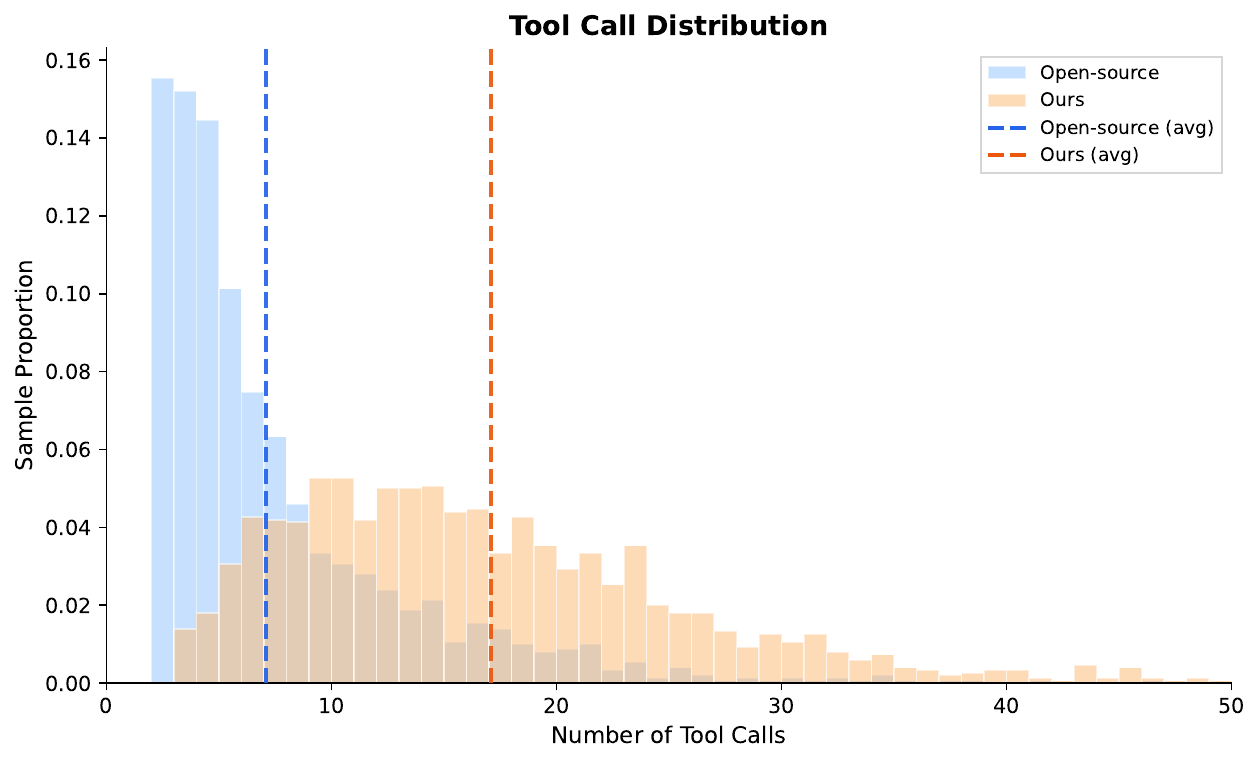}
    \end{minipage}
    \caption{Distribution comparison between the open-source multi-hop QA dataset (2Wiki, BeerQA, etc.) and our synthesized data. Left: token count per sample. Right: tool-call rounds per sample.}
    \label{fig:data-stats-length-tools-open-source}
\end{figure}

\begin{figure}[h]
    \centering
    \begin{minipage}[t]{0.49\linewidth}
        \centering
        \includegraphics[width=\linewidth]{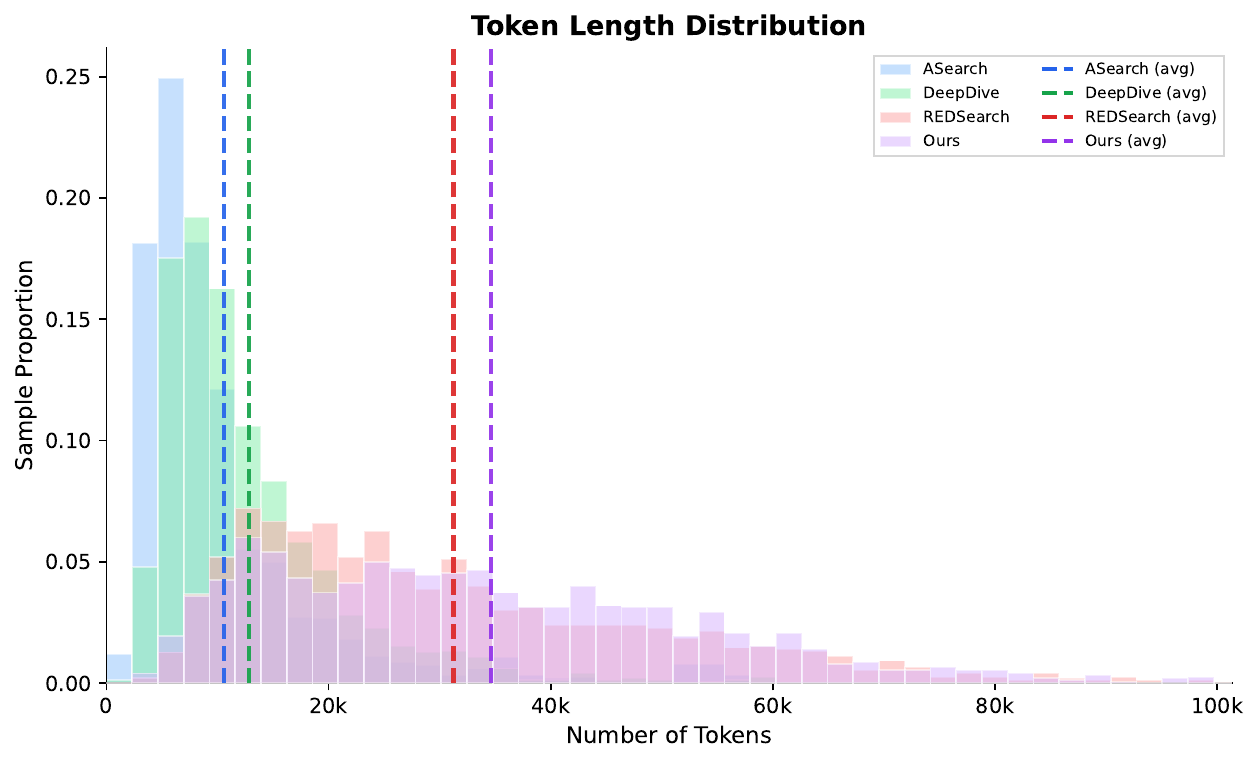}
    \end{minipage}
    \hfill
    \begin{minipage}[t]{0.49\linewidth}
        \centering
        \includegraphics[width=\linewidth]{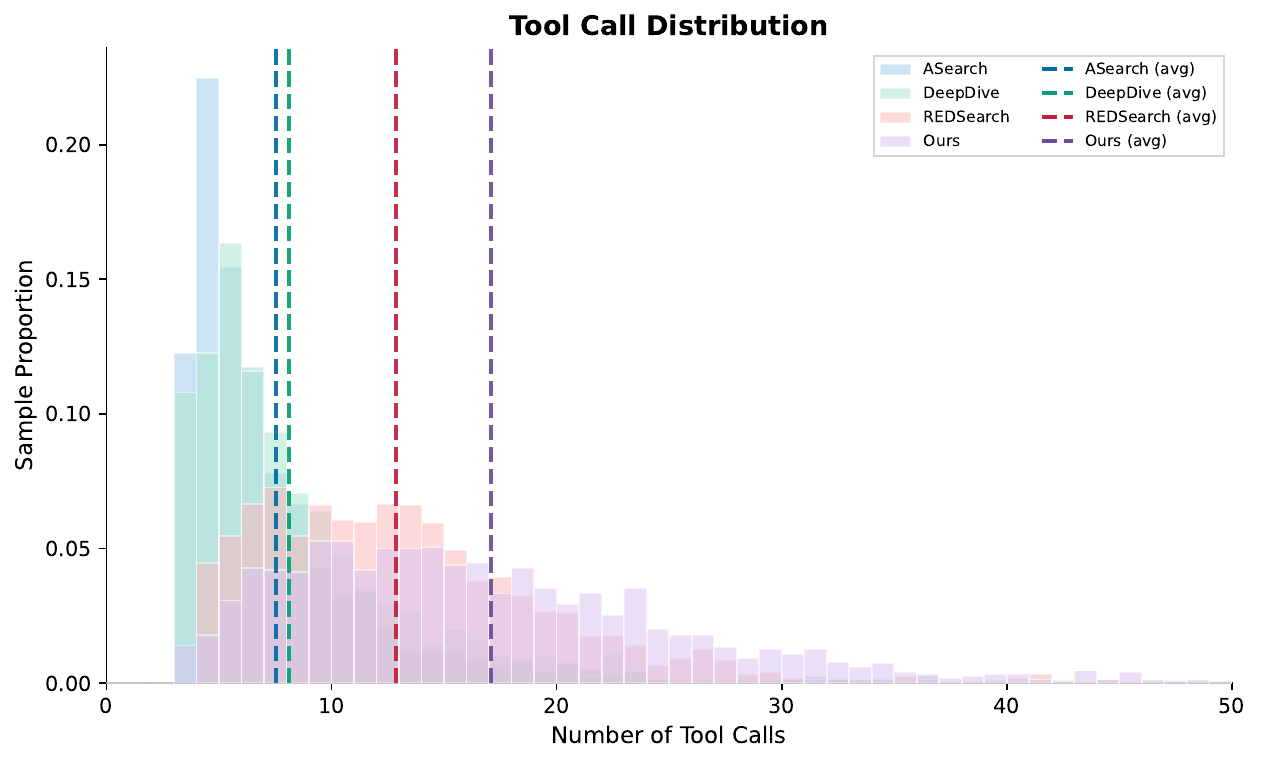}
    \end{minipage}
    \caption{Distribution comparison between three deep search data (ASearcher, DeepDive, and REDSearcher) and our synthesized data. Left: token count per sample. Right: tool-call rounds per sample. Our data is consistently shifted toward longer outputs and more tool interactions.}
    \label{fig:data-stats-length-tools}
\end{figure}

To analyze the strength of our constructed training dataset, we compare our synthesized data with three representative deep search datasets (REDSearcher~\cite{chu2026redsearcherscalablecostefficientframework}, DeepDive~\cite{lu2025deepdiveadvancingdeepsearch} and ASearcher~\cite{gao2025turnsunlockinglonghorizonagentic}) from two perspectives\footnote{The expert trajectory data corresponding to these QA datasets are all constructed using the same agent.}: token length and tool-use depth. Figure~\ref{fig:data-stats-length-tools-open-source} and Figure~\ref{fig:data-stats-length-tools} show that our synthesized samples have both longer token sequences and more tool-call rounds than existing multi-hop and deep search open-source dataset. This shift is important for deep-search training: longer trajectories provide denser supervision on cross-step reasoning, and deeper tool interaction exposes the model to more realistic long-horizon decision patterns. As a result, the model can better learn to maintain state, revise intermediate hypotheses, and complete complex tasks that require extended evidence aggregation.

\begin{figure}[h]
    \centering
    \includegraphics[width=0.75\linewidth]{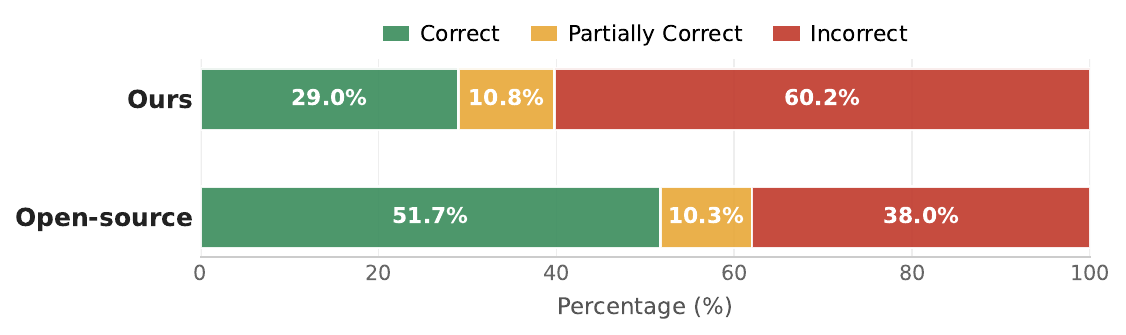}
    \caption{Relative to open-source data, our synthesized datasets indicate higher intrinsic difficulty.}
    \label{fig:judge-compare}
\end{figure}

Moreover, Figure~\ref{fig:judge-compare} further supports the difficulty of our data. Under the same ReAct-style trajectory construction methods using the same frontier agent, our generated data shows a lower answerable rate than open-source data (29.0\% < 51.7\%), which indicates a harder distribution.

\paragraph{Effect of QA data verification.}
To isolate the impact of our adversarial uniqueness verification on data quality, we compare two graph-based synthesis pipelines under identical data scales: one with verification and a baseline without.\footnote{We conduct this ablation using the graph-based approach due to its lower computational overhead. In contrast, agent-based synthesis requires multi-turn web exploration, making data construction prohibitively expensive.} As shown in Table \ref{tab:qa-verification}, integrating this verification step improves downstream performance across most benchmarks. By filtering out noisy and ambiguous samples, verification yields cleaner and more reliable data for subsequent trajectory construction and training.

\begin{table}[h]
    \centering
    \caption{Effect of adversarial uniqueness verification on graph-based QA data quality. Both rows use the same number of QA samples; the difference is whether verification is applied during synthesis. BC-200-sample is a sub-set of BrowseComp with 200 random samples.}
    \label{tab:qa-verification}
    \small
    \begin{tabular}{l c c c c}
    \toprule
    \textbf{QA Synthesis Method} & \textbf{BC-200-sample} & \textbf{BC-ZH} & \textbf{GAIA} & \textbf{xBench-DS-2505} \\
    \midrule
    Graph-based QA (w/o verification) & 14.2 & 24.5 & 55.3 & 67.0 \\
    Graph-based QA (w/ verification)  & 13.8 & 26.8 & 57.6 & 68.3 \\
    \quad \textcolor{red}{$\Delta$ Improvement} & \textcolor{blue}{-0.4} & \textcolor{red}{+2.3} & \textcolor{red}{+1.7} & \textcolor{red}{+1.3} \\
    \bottomrule
    \end{tabular}
\end{table}

\paragraph{Ablation Study on Verification-Driven Trajectory Construction.}

We evaluate whether adding multi-agent trajectories with explicit verification patterns improves performance. Concretely, we compare: (1)~single-agent ReAct trajectories only; and (2)~single-agent trajectories augmented with multi-agent verified trajectories.
Table~\ref{tab:framework-comparison} shows that augmented single-agent ReAct trajectories consistently improves performance across all benchmarks, with $+2.03$ average improvement. These results valid the contributions of trajectories with verification pattern.

\begin{table}[h]
    \centering
    \caption{Ablation study on verification-driven trajectory construction.}
    \label{tab:framework-comparison}
    \small
    \setlength{\tabcolsep}{3.5pt}
    \begin{adjustbox}{max width=\textwidth}
    \begin{tabular}{l c c c c}
    \toprule
    \textbf{Trajectory Source} & \textbf{BC-200-sample}& \textbf{BC-ZH} & \textbf{GAIA} & \textbf{xBench-DS-2505} \\
    \midrule
    Single-agent ReAct only & 13.8 & 26.8 & 57.6 & 68.3 \\
    Single-agent + Multi-agent (verified) & 14.5 & 27.0 & 62.8 & 70.3 \\
    \quad \textcolor{red}{$\Delta$ Improvement} & \textcolor{red}{+0.7} & \textcolor{red}{+0.2} & \textcolor{red}{+5.2} & \textcolor{red}{+2.0} \\
    \bottomrule
    \end{tabular}
    \end{adjustbox}
\end{table}

\paragraph{Improvement of Reinforcement Learning.}

We compare the SFT checkpoint with its RL-updated counterpart under the same evaluation setup to verify the effectiveness of the RL stage. Table~\ref{tab:rl-ablation} shows consistent gains from RL across all five benchmarks. Improvements range from $+0.8$ to $+6.7$ points, with an average gain of $+2.6$ points. This confirms that RL training on our constructed challenging QA data provides robust additional optimization on top of SFT.

\begin{table}[h]
    \centering
    \caption{Contribution brought from RL training on deep research benchmarks.}
    \label{tab:rl-ablation}
    \small
    \begin{tabular}{l c c c c}
    \toprule
    \textbf{Training Stage} & \textbf{GAIA} & \textbf{xBench-DS-2505} & \textbf{BC-200-sample} & \textbf{BC-ZH} \\
    \midrule
    Marco DeepResearch (8B) SFT & 59.2 & 68.3 & 16.5 & 27.1 \\
    Marco DeepResearch (8B) RL & 61.2 & 75.0 & 17.3 & 29.3 \\
    \quad \textcolor{red}{$\Delta$ Improvement} & \textcolor{red}{+2.0} & \textcolor{red}{+6.7} & \textcolor{red}{+0.8} & \textcolor{red}{+2.2} \\
    \bottomrule
    \end{tabular}
\end{table}

\paragraph{Test-Time Scaling.}

We evaluate the effectiveness of our proposed test-time scaling strategy on the top of the RL checkpoint. Table~\ref{tab:inference-comparison} shows that our strategy delivers substantial gains at test time. Compared with the RL baseline, performance improves by $+8.7$ on GAIA, $+7.0$ on xBench-DeepSearch-2505, $+15.0$ on BrowseComp-200-sample, and $+17.8$ on BrowseComp-ZH. The average gain is $+12.1$ points, indicating the potential of our proposed test-time scaling strategy.

\begin{table}[h]
    \centering
    \caption{Contribution of our proposed verifier-guided test-time scaling.}
    \label{tab:inference-comparison}
    \small
    \setlength{\tabcolsep}{3.5pt}
    \begin{adjustbox}{max width=\textwidth}
    \begin{tabular}{l c c c c}
    \toprule
    \textbf{Inference Strategy} & \textbf{GAIA} & \textbf{xBench-DS-2505} & \textbf{BC-200-sample} & \textbf{BC-ZH} \\
    \midrule
    Marco DR (8B) SFT+RL & 61.2 & 75.0 & 17.3 & 29.3 \\
    + Discard-all & 61.5 & 72.0 & 23.7 & 38.9 \\
    + Discard-all + Verify & 69.9 & 82.0 & 32.3 & 47.1 \\
    \quad \textcolor{red}{$\Delta$ vs.\ baseline} & \textcolor{red}{+8.7} & \textcolor{red}{+7.0} & \textcolor{red}{+15.0} & \textcolor{red}{+17.8} \\
    \bottomrule
    \end{tabular}
    \end{adjustbox}
\end{table}

\paragraph{Context Window Extension during Training.}

We further study whether extending the training context window improves long-horizon deep-search performance. Using the same SFT setup, data, and evaluation protocol, we compare 64K and 128K context settings. As shown in Table~\ref{tab:context-length}, increasing the context window from 64K to 128K yields consistent gains on both benchmarks, with improvements of $+2.3$ on BrowseComp-200-sample and $+0.8$ on BrowseComp-ZH (average: $+1.6$). This result supports the importance of long-context training for deep-search tasks that require many tool calls and cross-page evidence aggregation.

\begin{table}[h]
    \centering
    \caption{Effect of extending SFT context length from 64K to 128K.}
    \label{tab:context-length}
    \small
    \begin{tabular}{l c c}
    \toprule
    \textbf{Models} & \textbf{BC-200-sample} & \textbf{BrowseComp-ZH} \\
    \midrule
    Marco DeepResearch (8B) SFT 64K & 14.2 & 26.3 \\
    Marco DeepResearch (8B) SFT 128K & 16.5 & 27.1 \\
    \quad \textcolor{red}{$\Delta$ Improvement} & \textcolor{red}{+2.3} & \textcolor{red}{+0.8} \\
    \bottomrule
    \end{tabular}
\end{table}



\section{Conclusion}
\label{sec:conclusion}

This paper addresses a bottleneck in current deep research agents: the lack of explicit verification across QA data synthesis, trajectory construction, and inference, which leads to error propagation and under-utilized test-time computation. To solve this, we propose Marco DeepResearch, a 8B-scale deep search agent that optimized by our proposed verification-centric design with three improvements: verified QA synthesis, verification-driven trajectory construction, and verifier-guided test-time scaling. Extensive experimental results demonstrate that our proposed Marco DeepResearch agent significantly outperforms 8B-scale open-source deep search agents on most benchmarks BrowseComp and BrowseComp-ZH, and surpasses several 30B-scale deep search agents on BrowseComp-ZH. Furthermore, detailed analysis and ablation study verify the positive contributions of our verification-centric designs.

\bibliography{our}

\appendix
\newpage
\section*{Contributions and Acknowledgements}
\label{sec:appendix-contributors}
The development of Marco DeepResearch is a highly collaborative effort involving all members of our team.

\paragraph{Project Lead.} Longyue Wang.

\paragraph{Core Contributors.} Bin Zhu, Qianghuai Jia, Tian Lan, Junyang Ren, Feng Gu, Feihu Jiang, Longyue Wang, Zhao Xu, Weihua Luo.

\paragraph{Contributors.} Yu Zhao, E. Zhao, Jingzhen Ding, Yuxuan Han, ChenLin Yao, Jianshan Zhao, Wanying Chen, Jiahong Wang, Jiahe Sun, Wanghui huang, Yongchao Ding, Junyuan Luo, Junke Tang, Zhixing Du, Zhiqiang Yang, Haijun Li, Huping Ding.

\end{document}